\documentclass[letterpaper, 10 pt, journal, twoside]{ieeetran}

\IEEEoverridecommandlockouts

\usepackage{graphicx, import}
\usepackage{amsmath} 
\usepackage{amssymb} 

\usepackage{afterpage}
\usepackage{subcaption}
\usepackage{float}
\usepackage{bibentry}

\let\subparagraph\paragraph
\usepackage{titlesec}

\usepackage{booktabs}
\usepackage{multirow}
\usepackage{algorithmic}
\usepackage{algorithm}
\usepackage{calc}
\usepackage[usenames, dvipsnames]{xcolor}

\usepackage[acronyms, shortcuts]{glossaries}

\makeatletter
\let\NAT@parse\undefined

\newcommand\fs@betterruled{%
  \def\@fs@cfont{\bfseries}\let\@fs@capt\floatc@ruled
  \def\@fs@pre{\vspace*{5pt}\hrule height.8pt depth0pt \kern2pt}%
  \def\@fs@post{\kern2pt\hrule\relax}%
  \def\@fs@mid{\kern2pt\hrule\kern2pt}%
  \let\@fs@iftopcapt\iftrue}
\floatstyle{betterruled}
\restylefloat{algorithm}
\makeatother

\usepackage[breaklinks=true,colorlinks,bookmarks=false]{hyperref}
\usepackage[noabbrev,capitalise]{cleveref}

\usepackage{url}
\usepackage{color}
\usepackage[colorinlistoftodos]{todonotes}

\titlespacing*{\section}{0pt}{1.5mm}{1.5mm}
\titlespacing*{\subsection}{0pt}{.75mm}{.75mm}
\titlespacing*{\subsubsection}{0pt}{1mm}{1mm}

\usepackage{etoolbox}
\newtoggle{ext}
\toggletrue{ext}

\iftoggle{ext}{}{\renewcommand{\baselinestretch}{0.98}}
\newlength{\oldtextfloatsep}

\newacronym[longplural={mixed-integer convex programs}]{micp}{MICP}{mixed-integer convex program}

\newacronym[longplural={second-order cone programs}]{socp}{SOCP}{second-order cone program}
\newacronym[longplural={mixed-integer second-order cone programs}]{misocp}{MISOCP}{mixed-integer second-order cone program}

\newacronym[longplural={quadratically constrained convex programs}]{qcqp}{QCQP}{quadratically constrained convex program}
\newacronym[longplural={mixed-integer quadratically constrained convex programs}]{miqcqp}{MIQCQP}{mixed-integer quadratically constrained quadratic program}

\newacronym[longplural={quadratic programs}]{qp}{QP}{quadratic program}
\newacronym[longplural={sequential quadratic programs}]{sqp}{SQP}{sequential quadratic program}
\newacronym[longplural={mixed-integer quadratic programs}]{miqp}{MIQP}{mixed-integer quadratic program}

\newacronym{mpc}{MPC}{model predictive control}
\newacronym{pwa}{PWA}{piecewise affine}

\newacronym{mlopt}{MLOPT}{Machine Learning Optimizer}

\newcommand{\eg}{{e.g.}}
\newcommand{\ie}{{i.e.}}

\newcommand{\coco}{CoCo}

\newcommand{\reals}{{\mathbf{R}}}

\newcommand{\bin}{\delta}
\newcommand\mydots{\makebox[1em][c]{.\hfil.\hfil.}}

\newcommand*{\refeq}[1]{%
  \begingroup
    \hypersetup{
      linkcolor=black,
      linkbordercolor=black,
    }%
    \ref{#1}%
  \endgroup
}   

\title{\LARGE \bf
CoCo: Online Mixed-Integer Control via Supervised Learning
}

\author{
\parbox{\linewidth}{\centering
      Abhishek Cauligi$^1$, Preston Culbertson$^2$, Edward Schmerling$^1$\\
 Mac Schwager$^1$, Bartolomeo Stellato$^3$, and Marco Pavone$^1$
    }%

\thanks{$^1$Department of Aeronautics and Astronautics, Stanford University.
        {\tt\small  \{acauligi, schmrlng, schwager, pavone\}@stanford.edu}}
\thanks{$^2$Department of Mechanical Engineering, Stanford University. {\tt\small  pculbertson@stanford.edu}}
\thanks{$^3$Department of Operations Research and Financial Engineering, Princeton University. {\tt\small  bstellato@princeton.edu}}
\thanks{This work was supported in part by NASA under the NASA Space Tech-nology Research Fellowship Grants NNX16AM78H and 80NSSC18K1180.}
}

\begin{document}

\maketitle
\thispagestyle{empty}
\pagestyle{empty}

\begin{abstract}
Many robotics problems, from robot motion planning to object manipulation, can be modeled as \acp{micp}. However, state-of-the-art algorithms are still unable to solve \acp{micp} for control problems quickly enough for online use and existing heuristics can typically only find suboptimal solutions that might degrade robot performance. In this work, we turn to data-driven methods and present the Combinatorial Offline, Convex Online (\coco{}) algorithm for quickly finding high quality solutions for \acp{micp}. \coco{} consists of a two-stage approach. In the offline phase, we train a neural network classifier that maps the problem parameters to a {\it logical strategy}, which we define as the discrete arguments and relaxed big-M constraints associated with the optimal solution for that problem. Online, the classifier is applied to select a candidate logical strategy given new problem parameters; applying this logical strategy allows us to solve the original \ac{micp} as a convex optimization problem. We show through numerical experiments how~\coco{} finds near optimal solutions to \acp{micp} arising in robot planning and control with 1 to 2 orders of magnitude solution speedup compared to other data-driven approaches and solvers.
\end{abstract}

\section{Introduction}\label{sec:intro}
Planning and control using \ac{micp} has been an extensive area of study within the robotics community. \acp{micp} can be used as a modeling framework to capture the rich set of behaviors and  logical constraints that arise in problems such as planning for systems with contact~\cite{DeitsKoolenEtAl2019,MarcucciTedrake2020}, motion planning~\cite{LandryDeitsEtAl2016,CulbertsonBandyopadhyayEtAl2019}, and dexterous manipulation~\cite{HoganGrauEtAl2018}. Despite their popularity, \acp{micp} have rarely been put into practice for real-world control tasks with demanding performance requirements of 10-100Hz operational rates due to computational constraints. Indeed, the tremendous strides made in accelerating \ac{micp} solution times by several orders of magnitude in the past few decades rely on multithreaded implementations that are inapplicable on embedded systems commonly found on robot hardware.
Thus, although algorithms such as branch-and-bound~\cite{LeeLeyffer2012} provide certificates of optimality for \acp{micp}, finding the optimal solution can be challenging in practice due to the $\mathcal{NP}$-hard nature of solving \acp{micp}. 

\begin{figure}[t!]
\centering
\includegraphics[width=0.93\columnwidth]{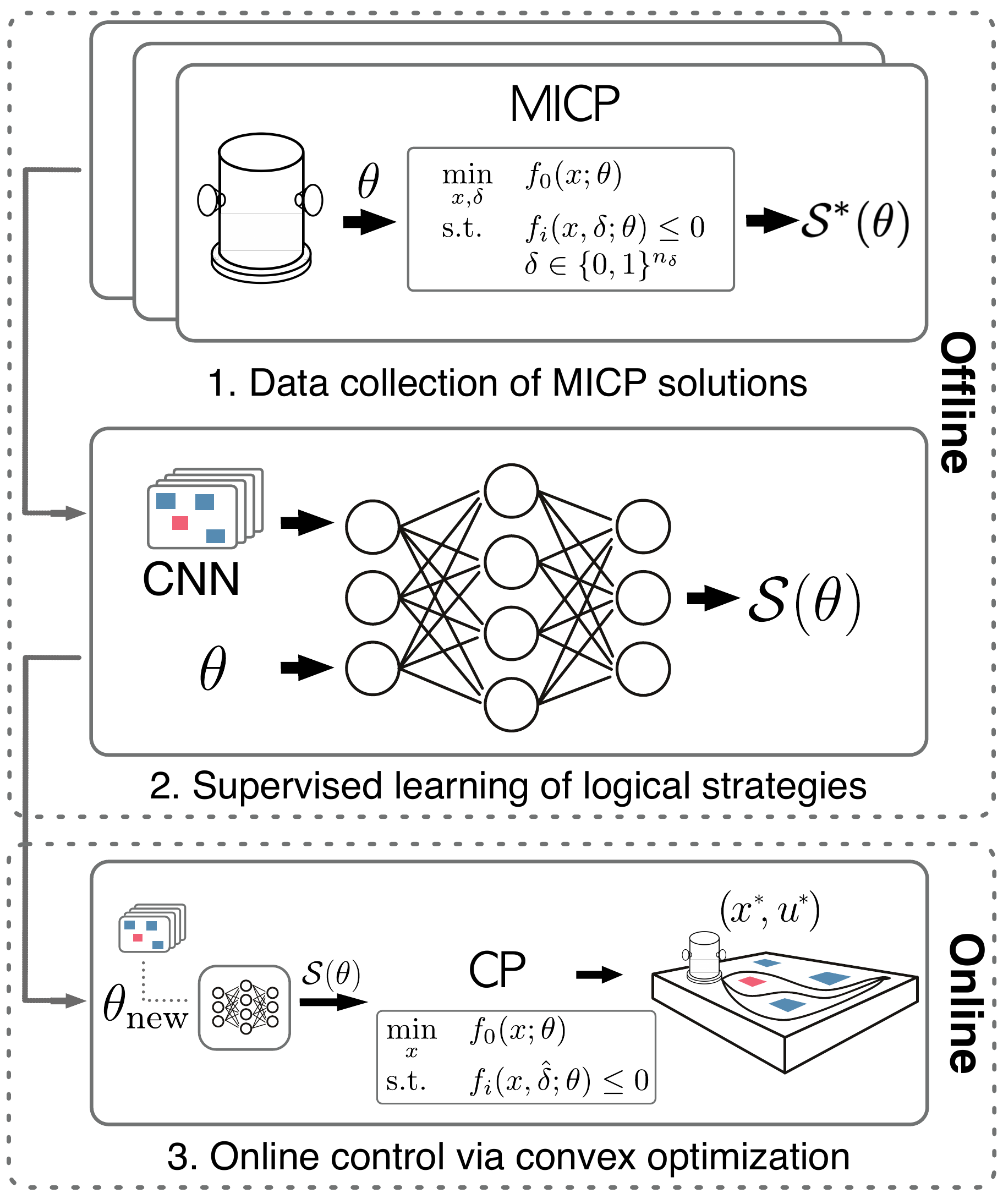}
\caption{%
Our algorithm~\coco{} is a data-driven approach that seeks to accelerate finding high-quality solutions to parametrized~\acp{micp}. The approach proposes a logical strategy $\hat{\mathcal{S}}(\theta)$, which is a candidate discrete solution $\hat{\bin}$ satisfying the logical constraints of the system. Given $\hat{\delta}$, the~\ac{micp} can be approximately solved as a convex optimization problem. Here, we show~\coco{} applied to the free-flying spacecraft robot motion planning problem. (1) Offline,~\coco{} solves a set of \acp{micp} for a representative set of planning problems and constructs the optimal logical strategy $\mathcal{S}^*(\theta)$ using the discrete optimizer $\bin^*$ and set of {\it relaxed} constraints $\mathcal{T}_M (\theta)$. (2) Thereafter, a convolutional neural network classifier is trained to learn a mapping between problem parameters $\theta$ and the logical strategy $\mathcal{S}^*(\theta)$. (3) Online, this classifier predicts the logical strategy $\hat{\mathcal{S}}(\theta)$ associated with new parameters $\theta$ and uses the candidate discrete solution $\hat{\bin}$ to solve convex optimization problems until a feasible solution is found.
}
\label{fig:free-flyer_table}
\end{figure}

Alternate techniques used to make \acp{micp} amenable for real-time control include terminating branch-and-bound when a feasible solution is first found or by simply rounding fractional integer solutions. However, such methods can degrade robot performance if a poor quality feasible solution is returned. A promising approach that has emerged in recent years is to apply techniques from machine learning to accelerate finding solutions for numerical-optimization based robot controllers~\cite{ZhangBujarbaruahEtAl2019}. Although melding supervised learning techniques and \ac{micp}-control has been considered~\cite{MastiBemporad2019,ZhuMartius2019}, these approaches have yet to demonstrate the ability to scale to the large number of discrete decision variables or problem parameters typically found in robotics. Further, these approaches are agnostic to how the discrete decision variables are utilized (\eg{}, piecewise affine constraints, mixed logical dynamics) and do not take into consideration the logical structure of the problem in their solution approach.


In this work, we seek to develop a learning-based methodology for applying \ac{micp}-based control with the following desiderata in mind:
\begin{enumerate}
    \item \emph{Performant control: } The controller should be able to find high quality solutions with respect to some performance metric for the control task.
    \item \emph{Speed: } The online solution procedure should be capable of providing real-time decision making.
    \item \emph{Generalizability: } Discrete decision variables are used to encode a rich set of behaviors in robotics and we seek an approach that can leverage the underlying structure present in many robot control tasks.
    \item \emph{Scalability: } The approach should be capable of solving \acp{micp} with a high dimensional parameter space and 10s-100s of discrete decision variables.
\end{enumerate}

{\em Related work:} The use of data-driven methods for quickly solving numerical optimization-based controllers is a nascent area of research.
In~\iftoggle{ext}{\cite{ZhangBujarbaruahEtAl2019,ChenWangEtAl2019}}{\cite{ZhangBujarbaruahEtAl2019}}, supervised learning is used for learning warm starts for a \ac{qp}-based controller.
Non-convex optimization-based controllers are considered as well, with~\iftoggle{ext}{\cite{WangTaubnerEtAl2020,TangSunEtAl2018}}{\cite{WangTaubnerEtAl2020}}  learning warm starts for a \ac{sqp}-based trajectory optimization library. The authors in~\cite{AgrawalBarrattEtAl2019b} leverage differentiable convex optimization to develop a learning-based approach for tuning a \ac{qp}-controller.
However, the shortcoming of these preceding approaches is that they are limited to continuous optimization problems.

In comparison, there is a paucity of approaches that have investigated the use of data-driven techniques for accelerating integer program solutions~\iftoggle{ext}{in control~\cite{BengioLodiEtAl2018}.}{in control.}
For general discrete optimization problems, a popular approach has been to consider branch-and-bound as a sequential decision making problem and apply reinforcement learning approaches~\cite{DaiKhalilEtAl2017}.
The shortcoming of such techniques for control is that they still rely on solving branch-and-bound online and often require multiple neural network forward passes at each node.

More recently, both~\cite{MastiBemporad2019,HoganGrauEtAl2018} propose a supervised learning approach using a neural network to warm start the binary variable assignments for an \ac{miqp} controller and a k-nearest neighbors approach is proposed in~\cite{ZhuMartius2019}. In these approaches, the supervised framework maps problem parameters to a candidate discrete solution and, by fixing the discrete decision variables to the candidate solution values, the \ac{miqp} is solved as a \ac{qp}. However, these approaches do not address scalability to a high dimensional parameter space or large number of discrete decision variables. Further, their solution strategy of directly proposing a candidate discrete solution without considering how the variables are employed in the~\ac{micp} limits their applicability to higher dimensional parametric programs often found in robotics. Our proposed approach accommodates a broad set of systems for which \acp{micp} are used as a modeling tool, such as mixed logical dynamical systems and \ac{pwa} systems.


Our work draws inspiration from the field of explicit \ac{mpc}~\cite{MalyutaAcikmese2019}. Unlike standard multiparametric optimization approaches that learn a solution map corresponding to polyhedral partitions of the parameter space, our approach learns strategies for regions of a nonlinear transformation (learned via a neural network) of the parameter space of the problem.

{\em Statement of Contributions:} Towards filling the gaps in existing works, we present the Combinatorial Offline, Convex Online (\coco{}) framework. \coco{} is a data-driven framework for solving \acp{micp} and entails a two stage approach (as detailed in~\cref{fig:free-flyer_table}) consisting of an offline and online phase. We introduce the concept of logical strategies and show how they enable for~\coco{} to find high quality solutions for a broad class of problems modeled as \acp{micp}, including problems with 100s of binary decision variables and a large input parameter dimension.

To the best of our knowledge, \coco{} uniquely satisfies the four aforementioned desiderata. This paper extends a conference publication~\cite{CauligiCulbertsonEtAl2020} and provides the following additional contributions:
\begin{enumerate}
    \item Demonstration of how the notion of task-specific logical strategies can be exploited to solve problems with a varying number of discrete variables.
    \item Additional numerical results to compare~\coco{} against a commercial \ac{micp} solver and benchmark data-driven methods used to find feasible solutions for \acp{micp}.
    \item A thorough analysis of how~\coco{} can be deployed in practical situations and used across a variety of tasks.
\end{enumerate}

\section{Technical Background}
This section introduces the parametrized \acp{micp} studied in this work, the big-M constraint formulation approach, and the concept of well-posedness of \ac{micp} solutions.

\subsection{Parametrized \acp{micp}}
This work considers discrete optimization problems of the specific form known as parametrized mixed-integer convex programs. Given problem parameters $\theta$, we can define the following parametrized \ac{micp} with continuous decision variables $x\in\reals^{n_x}$, and binary decision variables $\bin \in \{0,1\}^{n_\bin}$:
\begin{equation} \label{eq:bcp}
\begin{array}{ll}
\underset{x,\bin}{\textrm{minimize}} \!\!\!&f_0(x;\theta) \\
\text{subject to}\!\!\!& f_i(x,\bin;\theta) \leq 0, \quad i = 1,\dots,m_f \\
& h_i(\bin; \theta) \leq 0, \quad i = 1, \dots, m_I\\
&\bin\in \{0,1\}^{n_\bin}.\\
\end{array}
\end{equation}
The objective function $f_0$ and inequality constraints $f_i$ are assumed convex with respect to $x$. The purely integer constraints $h_i$ are assumed linear with respect to $\bin$.

We note here that the binary decision variables $\bin$ are the ``complicating'' variables in the sense that~\eqref{eq:bcp} becomes much easier to solve if $\bin$ are temporarily held fixed and the resulting convex program solved in terms of $x$. Indeed, if an optimal discrete solution $\bin^*$ for~\eqref{eq:bcp} is provided, then the continuous optimizer $x^*$ for the \ac{micp} can be easily found by solving a single convex optimization problem,
\begin{equation*}
\begin{array}{ll}
\underset{x}{\textrm{minimize}} & f_0(x;\theta) \\
\text{subject to} & f_i(x,\bin^*;\theta) \leq 0, \:\: i = 1,\dots,m_f.\\
\end{array}
\end{equation*}
Thus, we see that identifying an optimizer $\bin^*$ for~\eqref{eq:bcp} allows a user to quickly find the optimal solution for an \ac{micp} and circumvent, e.g., exploring a full branch-and-bound tree.
To this end, a host of supervised learning approaches from\iftoggle{ext}{ \cite{MastiBemporad2019,ZhuMartius2019,BertsimasStellato2019,KargLucia2018,LoehrKlaucoEtAl2020}}{ \cite{BertsimasStellato2019,MastiBemporad2019,ZhuMartius2019}} utilize this insight to generate a candidate $\hat{\bin}$ given problem parameters $\theta$ for an \ac{micp}.

\subsection{Big-M Formulation}
In \acp{micp}, binary variables $\bin$ are often introduced in conjunction with what is known as big-M formulation to capture high-level discrete or logical behavior of the system. These big-M constraints then enforce the desired high-level behavior on the continuous variables $x$ for the robot task at hand, \eg{}, contact task assignment or hybrid control logic~\cite{BemporadMorari1999}.

As an example of a desired logical behavior, consider enforcing the constraint, 
\begin{equation}
f_1 (x;\theta) \leq 0 \ \lor  f_2 (x; \theta) \leq 0,
\label{eq:impl-constraint}
\end{equation}
where $\lor$ indicates an ``or'' relationship. The big-M method can be used to encapsulate this constraint by introducing an auxiliary term,
\begin{equation*}
M_i(\theta) := \sup_{x} f_i(x; \theta),\quad i = 1, 2
\end{equation*}
where $M_1 (\theta)$ is the least upper bound on the value attainable by $f_1 (x; \theta)$ and $M_2 (\theta)$ the least upper bound for $f_2 (x; \theta)$. Then, the logical behavior can be enforced through the introduction of two binary variables $\delta_1$ and $\delta_2$,
\begin{align}
f_1(x;\theta) \leq M_1(\theta) (1-\bin_1) \label{eq:eg_disjunctive_1}\\
f_2(x;\theta) \leq M_2(\theta) (1-\bin_2) \label{eq:eg_disjunctive_2}\\
\bin_1+\bin_2 \geq 1 \label{eq:eg_disjunctive_3}
\end{align}
The equivalence between~\eqref{eq:impl-constraint} and~\eqref{eq:eg_disjunctive_1}-\eqref{eq:eg_disjunctive_3} can be verified by noting that when $\bin_1=1$, then $f_1(x;\theta) \leq 0$ is automatically recovered. Similarly, $\bin_2 = 1$ leads to the constraint $f_2 (x;\theta) \leq 0$. Enforcing the behavioral constraint that one of the two constraints must be satisfied is accomplished through the final constraint~\eqref{eq:eg_disjunctive_3}.

We note that the right hand sides of~\eqref{eq:eg_disjunctive_1}-\eqref{eq:eg_disjunctive_2} are linear in terms of $\bin_i$ and thus denote a ``big-M'' constraint as an affine expression of $\bin_i$:
\begin{equation} \label{eq:big_M}
    g_i(x; \theta) \leq a_i(\theta) (1-\bin_i), 
\end{equation}
where $a_i(\theta)$ are positive-valued constants precomputed using $M (\theta)$, or any upper bound for the constraint, to impose the desired logical behavior.

In this work, we use big-M constraints exclusively to relate the continuous variables $x$ and binary variables $\bin$. If $m_M$ is the number of big-M constraints used, we can write more specifically the class of \acp{micp} studied as:
\begin{equation} \label{eq:bcp_logic}
\!\begin{array}{ll}
\underset{x,\bin}{\textrm{minimize}} & f_0(x;\theta) \\
\text{subject to} & f_i(x;\theta) \leq 0, \quad i = 1,\dots,m_f \\
& g_i(x; \theta) \leq a_i(\theta) (1-\bin_i), \quad i = 1,\dots,m_M\\
& h_i(\bin; \theta) \leq 0, \quad i = 1, \dots, m_I\\
&\bin\in \{0,1\}^{n_\bin}.\\
\end{array}
\tag{$\mathcal{P}(\theta)$}
\end{equation}

\subsection{Well-Posedness of \ac{micp} Solutions}
Here, we distinguish between two common classes of \acp{micp} and will later show how this difference is crucial in formulating an effective solution approach. Specifically, as the inclusion of binary variables $\bin$ in~\eqref{eq:bcp} renders the problem non-convex, an \ac{micp} may admit multiple globally optimal solutions ($x^*$, $\bin^*$). This leads to the distinction between \textit{well-posed} and \textit{completely well-posed} \acp{micp} as introduced in~\cite{BemporadMorari1999}. Well-posed \acp{micp} admit a unique continuous minimizer $x^*$, but may admit multiple discrete optimizers $\bin^*$. Completely well-posed problems assume that an \ac{micp} admits a single global minimizer ($x^*$, $\bin^*$). 

Completely well-posed \acp{micp} can be used to model many systems, such as those with \ac{pwa} constraints.
In the case of \ac{pwa} constraints, a continuous solution $x^*$ can only be attained by a particular set of mode transitions that are uniquely encoded by a single discrete optimizer $\bin^*$.
More broadly however, the assumption of a unique discrete optimizer is a limiting one and cannot accomodate many applications.
For example, mixed logical dynamical systems~\cite{BemporadMorari1999} enforce logic rules on discrete-time dynamical systems with both continuous and discrete decision variables, but allow for multiple discrete optimizers $\{ \bin^* \}$ that satisfy the propositional logic constraints.
Thus, we assume that the problems we treat are well-posed \acp{micp}.
Although this entails assuming that the \ac{micp} admits a single $x^*$, we note that this is a mild assumption in practice.
For example, if problem parameters $\theta$ yield multiple $x^*$ for~\ref{eq:bcp_logic}, a small perturbation in the initial condition typically breaks ties and leads to a single minimizer $x^*$.
Indeed, in many robotics problems, slight regularization terms can be added to the objective function to ensure a unique $x^*$ while still accomplishing the control task.

Given a continuous optimizer $x^*$, the set of binary optimizers $\{ \bin^* \}$ for a well-posed~\ac{micp} can be characterized by identifying which big-M constraints of the form~\eqref{eq:big_M} are {\it relaxed}, where a constraint $g_i (x^*; \theta)$ is denoted relaxed if:
\begin{equation} \label{eq:relaxed_big_M}
g_i(x^*; \theta) > 0.
\end{equation}
That is, a big-M constraint is considered relaxed if, after plugging in the values from $x^*$, the constraint~\eqref{eq:big_M} is satisfied if and only if the binary variable $\bin^*_i$ attains value $0$. %
\footnote{We stress here that a relaxed constraint does not connote a relaxation of the binary variable $\bin_i$ over its convex hull, $\bin_i \in [0,1]$.}
If $\bin$ is subject to purely integer constraints (\eg{}, a cardinality constraint), then the purely integer constraints are also included in identifying the list of relaxed constraints.
In contrast, a big-M constraint is considered {\it enforced} if the values for $x^*$ automatically lead to constraint satisfaction, \ie{}, $g_i(x^*; \theta) \leq 0$ which allows for both cases of $\bin_i^*=0$ or $\bin_i^*=1$ (subject to purely integer constraints).
Thus, the set $\{ \bin^* \}$ is then comprised of $\bin^*$ that attain value $\bin^*_i=0$ for the set of relaxed constraints and only differ in the values associated with the enforced constraints.

As an example, we consider a continuous optimizer $x^*$ that satisfies the logical expression from~\eqref{eq:impl-constraint} for which $f_1 (x^*;\theta) \leq 0$, but $f_2 (x^*;\theta) > 0$.
In this case, $f_2 (x^*; \theta)$ is a relaxed constraint, while $f_1 (x^*; \theta)$ is an enforced constraint.
This value of $x^*$ then admits two binary optimizers $(\bin_1^*, \bin_2^*)=(0,1)$ and $(\bin_1^*, \bin_2^*)=(1,1)$.
We see here that as $f_2 (x^*; \theta)$ is the relaxed constraint, $\bin_2^*$ attains value $1$ for both possible binary optimizers.

\section{Technical Approach}
This section defines logical strategies, presents the subsequent development of task-specific logical strategies, and demonstrates how they can be used for solving \acp{micp}.

\subsection{Logical Strategies}
\acp{micp} are used to encode a rich set of logical constraints and behaviors for robotic systems and existing approaches seek to tackle this broad class of problems by identifying an optimal discrete solution $\bin^*$ associated with problem parameters $\theta$. For well-posed \acp{micp}, a solution approach that only proposes a single candidate binary optimizer $\hat{\bin}$ given $\theta$ leads to an ill-posed supervised learning problem, as there may exist a set of target values $\{ \bin^* \}$ given a $\theta$.

To address this shortcoming, we present the notion of logical strategies, named as such because they take into consideration how binary variables $\bin$ are used to enforce high-level logical behavior in well-posed~\acp{micp}. A logical strategy $\mathcal{S}(\theta )$ leverages the idea that, while there may exist multiple discrete optimizers given a continuous optimizer $x^*$, the problem~\refeq{eq:bcp_logic} can be solved by identifying the set of relaxed big-M constraints.

Given the preceding definition of a relaxed big-M constraint from~\eqref{eq:relaxed_big_M}, we now define a logical strategy. Given problem parameters $\theta$ and continuous optimizer $x^*$ for the problem~\refeq{eq:bcp_logic}, let $\bin^*(\theta)$ be a particular binary optimizer from the set of discrete optimizers $\{  \bin^* (\theta) \}$ and $\mathcal{T}_M(\theta)$ be the set of relaxed big-M constraints for the problem,
\begin{align}
\mathcal{T}_M(\theta) = \{ i \mid g_i(x^*; \theta) > 0\}.
\end{align}
We then define the logical strategy $\mathcal{S}(\theta)$ as a tuple $(\bin^*(\theta), \mathcal{T}_M(\theta))$.

The utility of the logical strategy definition becomes apparent when revisiting the original \ac{micp} in~\eqref{eq:bcp}. If the optimal logical strategy $\mathcal{S}^* (\theta)$ is provided for an \ac{micp}, then the continuous optimizer $x^*$ for the \ac{micp} can be found by solving a single convex optimization problem,
\begin{equation*}
\begin{array}{ll}
\underset{x}{\textrm{minimize}} & f_0(x,\bin^*;\theta) \\
\text{subject to} & f_i(x;\theta) \leq 0, \quad i = 1,\dots,m_f \\
& g_i(x; \theta) \leq 0, \quad i \in \mathcal{T}_M^c(\theta)\\
\end{array}
\end{equation*}
where $\mathcal{T}_M^c(\theta)$ is the set of enforced constraints, \ie{}, the complement of $\mathcal{T}_M(\theta)$.
We note here that if~\ref{eq:bcp_logic} admits a set of discrete solutions $ \{ \bin^* \}$ given a particular $x^*$, then any one of the $\bin^*$ can be included in $\mathcal{S}(\theta)$. 
This leads to the insight that, rather than proposing a candidate $\hat{\bin}(\theta)$, a solution approach should rather propose a candidate logical strategy $\hat{\mathcal{S}}(\theta)$.
As we demonstrate later, considering logical strategies improves performance of a supervised learning-based solution approach.

\subsection{Task-Specific Logical Strategies for \acp{micp}}
For large problems, proposing a candidate logical strategy starts to become intractable as the number of candidate logical strategies becomes too large. However, underlying problem structure allows us to consider the logical strategy associated with each constraint individually. To that end, we introduce the idea of task-specific logical strategies that consider a common structure arising in many robotics problems known as separability.

To demonstrate an example of separability, we consider the robot motion planning problem shown in Figure~\ref{fig:free-flyer_table}. Four binary variables are used to enforce obstacle avoidance, where each binary variable is associated with lying on one side of an axis-aligned rectangular obstacle at a particular time. Thus, the binary variables $\bin$ are decoupled on the basis of which obstacle they are associated with.

Task-specific logical strategies seek to exploit this problem structure. Formally, the underlying mixed logical constraints can be written as a conjunction of Boolean formulas:
\begin{equation*}F_1 \land F_2 \land \mydots \land F_\ell,
\end{equation*}
where $F_i$ is a distinct sub-formula of literals involving continuous and binary variables, and each binary variable $\bin_j$ is associated with only one sub-formula $F_i$.
When the mixed logical constraint consists of $\ell$ such Boolean formulas, then the logical strategy $\mathcal{S} (\theta)$ can itself be split into $\ell$ sub-formula strategies $\mathcal{S}_1 (\theta), \mydots, \mathcal{S}_\ell (\theta)$.
We note that this decomposition is only possible because of the separability in the problem that allows for each $\bin_j$ to be considered only in relation to a single sub-formula $\mathcal{S}_i (\theta)$ and the value of the continuous solution $x^*$.

An example of such separability arising in a robotics problem is the~\ac{micp} formulation of collision avoidance constraints~\cite{LandryDeitsEtAl2016}.
Consider an axis-aligned, 2D rectangular obstacle $m$ that is parametrized by the coordinates of its lower-left hand corner $(x^m_\textrm{min}, y^m_\textrm{min})$ and upper right-hand corner $(x^m_\textrm{max}, y^m_\textrm{max})$.
If the 2D position of the robot is $p=(p^1, p^2)\in\reals^2$, then the collision avoidance constraints with respect to obstacle $m$ are:
\begin{align}
x_{\textrm{max}}^m - M\bin^{m,1} \leq p^1 \leq x_{\textrm{min}}^m + M\bin^{m,2} \label{eq:obstacle_avoidance_x}\\
y_{\textrm{max}}^m - M\bin^{m,3} \leq p^2 \leq y_{\textrm{min}}^m + M\bin^{m,4} \label{eq:obstacle_avoidance_y}
\end{align}
where $M$ is chosen to be a sufficiently large number.
As written in~\eqref{eq:obstacle_avoidance_x} and~\eqref{eq:obstacle_avoidance_y}, $\bin^{m,i}=1$ indicates that robot is on one side of face $i$ of the obstacle and in violation of that keep-out constraint.
To ensure that the robot does not collide with obstacle $m$, a final constraint
\begin{equation}
\begin{array}{l}
\sum\limits_{i=1}^4 \bin^{m,i} \leq 3, \label{eq:obstacle_avoidance_sum}
\end{array}
\end{equation}
is enforced. Note that each binary variable depends only on the three other variables associated with the same obstacle. Thus, each task-specific logical strategy considers only the binary variables and big-M constraints for obstacle $m$.

Task-specific logical strategies offer several advantages for a supervised learning approach to solving parametrized \acp{micp}.
First, as the number of possible logical strategies can grow exponentially in terms of the number of binary variables $n_\bin$, considering each sub-formula strategy separately $\mathcal{S}_i (\theta)$ reduces the number of binary variables in each task-specific logical strategy, thereby resulting in a smaller number of values that each sub-formula can attain.
Second, this reduced number of candidate sub-formulas leads to improved supervision in a learning-based approach as the number of target values is reduced.
Finally, in the case when additional Boolean formulas are added (\ie{}, additional binary variables are added to the \ac{micp}), the sub-formula strategies can be queried at inference time for these new formulas.

Thus, task-specific logical strategies can lead to improved performance in problems with separability by reducing the number of class labels and thereby improving supervision for a data-driven approach. However, as we show, the separability of the constraints  has performance limits, especially in applications where additional sub-formulas strategies are queried at test time. Practically, in the context of robotics, the performance of the controller can be assessed beforehand and only deployed for tasks that are sufficiently similar to the test set (\eg{}, a maximum number of obstacles for which task-specific logical strategies are queried).

\section{Combinatorial Offline, Convex Online}
We now present our proposed approach \coco{}, short for Combinatorial Offline, Convex Online. \coco{} consists of a two-stage approach for training and deploying a neural network classifier that maps problem parameters $\theta$ to a candidate logical strategy $\hat{\mathcal{S}}(\theta)$.

Algorithm~\ref{alg:COCO_OFFLINE} details the offline portion of~\coco{}. The algorithm takes as input a set of problem parameters $\{ \theta_i \}$, where each $\theta_i$ is sampled from a parameter distribution $p (\Theta)$ representative of the problems encountered in practice. We refer to $\mathcal{S}$ as the strategy dictionary and $\mathcal{D}$ as the train set, both of which are initially empty (Line~\ref{line:init_dicts}). The strategy dictionary $\mathcal{S}$ stores the set of logical strategies $\{ \mathcal{S}^{(i)} \}$ constructed during the offline phase and the train set $\mathcal{D}$ stores the set training tuples $\{ (\theta_i, y_i) \}$ used for training the neural network classifier, where $y_i$ is the class label associated with logical strategy $\mathcal{S}^{(i)}$. For each $\theta_i$, the \ac{micp}~\refeq{eq:bcp_logic} is solved (Line~\ref{line:solve_micp}). If an optimal solution is found to the \ac{micp}, then the primal solution $(x^*, \bin^*)$ is used to construct the logical strategy $\mathcal{S}^*$ for this problem (Line~\ref{line:construct_strategy}) and added to the strategy dictionary $\mathcal{S}$ if it is not already included (Lines~\ref{line:add_strategy}-\ref{line:add_strategy_end}). The class label $y_i$ associated with $\mathcal{S}^*$ is identified (Line~\ref{line:assign_label}) and the tuple $(\theta_i, y_i)$ added to $\mathcal{D}$ (Line~\ref{line:add_sample}). Finally, a neural network classifier $\hat{h}_\phi$ with output dimension $| \mathcal{S} |$ is trained using the elements of $\mathcal{D}$ and the weights $\phi$ are chosen in order to approximately minimize a cross-entropy loss over the training samples (Line~\ref{line:nn_training}).


\setlength{\textfloatsep}{0.12cm}
\setlength{\floatsep}{0.12cm}

\begin{algorithm}[t]
\caption{\coco{} Offline}
\label{alg:COCO_OFFLINE}
\begin{algorithmic}[1]
{\small
    \REQUIRE {Batch of training data} $\{ \theta_i\}_{i=1,\ldots ,T}${, problem}~\refeq{eq:bcp_logic}
    \STATE Initialize strategy dictionary $\mathcal{S} \leftarrow \{ \},$ {train set } $\mathcal{D} \leftarrow \{ \}$ \label{line:init_dicts}
    \STATE $k \leftarrow 0$
    \FOR{{each} $\theta_i$}
        \STATE Solve \refeq{eq:bcp_logic} \label{line:solve_micp}
        \IF{\refeq{eq:bcp_logic} is optimal}
            \STATE Construct optimal strategy $\mathcal{S}^*$ \label{line:construct_strategy}
            \IF{$\mathcal{S}^* \textbf{ not in } \mathcal{S}$} \label{line:add_strategy}
                \STATE Add $\mathcal{S}^*$ to $\mathcal{S}$
                \STATE Assign class label $y_k$ to $\mathcal{S}^*$ \label{line:add_strategy_class}
                \STATE $k \gets k + 1$
            \ENDIF \label{line:add_strategy_end}
            \STATE Identify class label $y_i$ for strategy class $\mathcal{S}^*$ \label{line:assign_label}
            \STATE Add $\left(\theta_i, y_i\right)$ to $\mathcal{D}$ \label{line:add_sample}
        \ENDIF
    \ENDFOR
    \STATE Choose network weights $\phi$ which minimize cross-entropy loss $\mathcal{L}({h}_\phi(\theta_i),y_i)_{i=1,\ldots,T}$ via stochastic gradient descent \label{line:nn_training}
    \RETURN ${h}_\phi$, $\mathcal{S}$
}
\end{algorithmic}
\end{algorithm}

\begin{algorithm}[t]
\caption{\coco{} Online}
\label{alg:COCO_ONLINE}
\begin{algorithmic}[1]
{\small
    \REQUIRE Problem parameters $\theta$, strategy dictionary $\mathcal{S}$, trained neural network ${h}_\phi$, $n_{\text{evals}}$
    \STATE Compute class scores ${h}_\phi (\theta)$ \label{line:nn_forward_pass}
    \STATE Identify top $n_{\text{evals}}$-scoring strategies in $\mathcal{S}$ \label{line:strategy_scoring}
    \FOR {$j = 1,\ldots,n_{\text{evals}}$}
        \IF {\refeq{eq:bcp_logic} is feasible for strategy $\mathcal{S}^{(j)}$} \label{line:solve_cp}
            \RETURN {Feasible solution $(x^*,\bin^*)$} \label{line:feas_soln}
        \ENDIF
    \ENDFOR
    \RETURN \text{failure} \label{line:return_fail}
}
\end{algorithmic}
\end{algorithm}

\section{Numerical Experiments}
In this section, we compare~\coco{} with commercial solvers and other data-driven approaches for solving \acp{micp}. We present results on three benchmark problems in robotics that are modeled as \acp{micp}: the control of an underactuated cart-pole with multiple contacts, dexterous manipulation for task-specific grasping, and the robot motion planning problem.

\subsection{Implementation Details}
For each system, we first generate a dataset by sampling $\theta$ from $p(\Theta),$ until a sufficient number of problems \refeq{eq:bcp_logic} are solved. For each system, we separate 90\% of the problems for training and the remaining 10\% for evaluation. For the cart-pole and dexterous manipulation problems, the neural network architecture consists of a standard ReLU feedforward network with three layers and 32 neurons per layer. For the free-flyer system, we used a CNN architecture with four convolutional layers followed by a feedforward network with three layers and 128 neurons per layer.

We implemented each example in Python and used the~\verb|PyTorch| machine learning library to implement our neural network models with the ADAM optimizer for training.
The \acp{micp} were written using the~\verb|cvxpy| modeling framework and solved using Mosek.
We further benchmark \coco{} against the commercial solver, the regression framework from~\cite{MastiBemporad2019}, and the k-nearest neighbors framework from~\cite{ZhuMartius2019}.
We disable presolve and multithreading to better approximate the computational resources of an embedded processor.
The network architecture chosen for the regressor was identical to the \coco{} classifier, updated with the appropriate number of integer outputs.
The code for our algorithm is available at~{\tt https://github.com/StanfordASL/CoCo}.

\subsection{Cart-Pole with Soft Walls}
\iftoggle{ext}{
\begin{figure}[t]
\centering
\def\svgwidth{0.6\columnwidth}
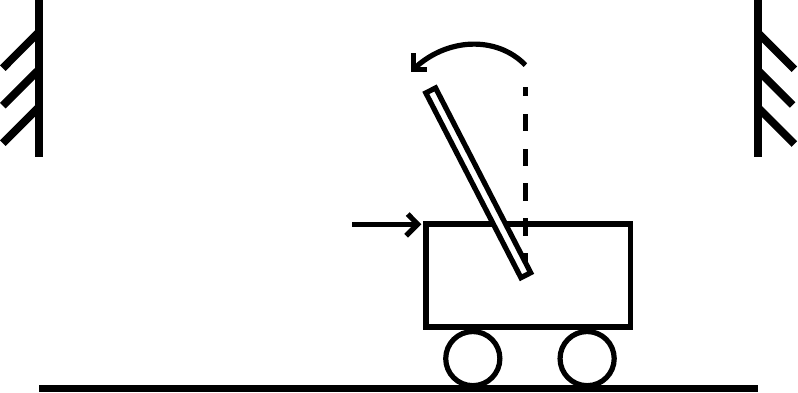
\caption{4D cart-pole with wall system.\\}
\label{fig:cart-pole_soft_walls}
\vspace{2em}
\end{figure}
}{}

\begin{figure*}[t!]
    \centering
    \hfill
    \begin{subfigure}[t]{0.23\textwidth}
        \includegraphics[width=\textwidth]{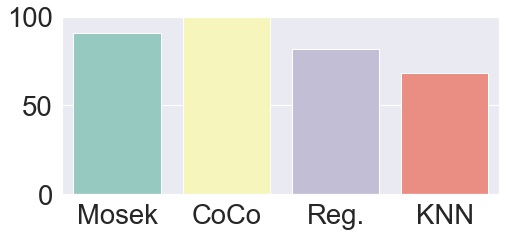}
        \caption{Success percentage}
        \label{fig:cart-pole_percent_succ}
    \end{subfigure}
    \hfill
    \begin{subfigure}[t]{0.23\textwidth}
        \includegraphics[height=0.75in,clip]{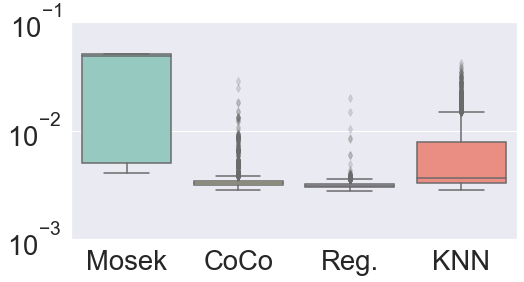}
        \caption{Solution times [s]}
        \label{fig:cart-pole_time}
    \end{subfigure}
    \hfill
    \begin{subfigure}[t]{0.23\textwidth}
        \includegraphics[height=0.75in,clip]{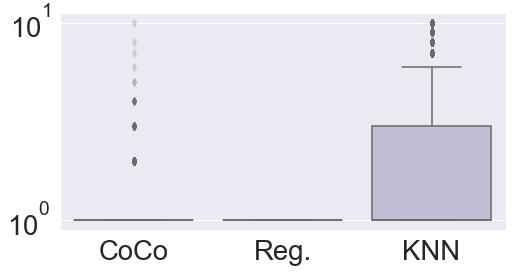}
        \caption{Num. \acp{qp} solved}
        \label{fig:cart-pole_solved}
    \end{subfigure}
    \hfill
    \begin{subfigure}[t]{0.23\textwidth}
        \includegraphics[width=\textwidth]{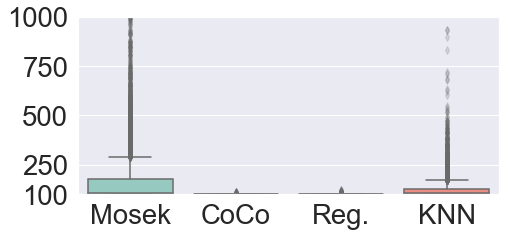}
        \caption{Normalized cost [\%]}
        \label{fig:cart-pole_cost}
    \end{subfigure}
    \hfill
    \vspace{1.5em}
  \caption{For the cart-pole system, \coco{} finds near-global solutions for a majority of problems.}
\label{fig:cart-pole_results}
\vspace{1.5em}
\end{figure*}

\begin{figure*}[t!]
    \centering
    \hfill
    \begin{subfigure}[t]{0.30\textwidth}
        \centering
        \includegraphics[height=0.7in,clip]{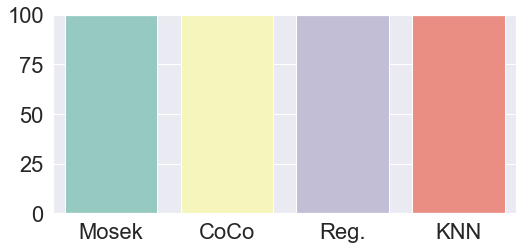}
        \vspace{-5pt}
        \caption{Success percentage}
        \label{fig:manipulation_percent_succ}
    \end{subfigure}
    \hfill
    \begin{subfigure}[t]{0.30\textwidth}
        \centering
        \includegraphics[height=0.7in,clip]{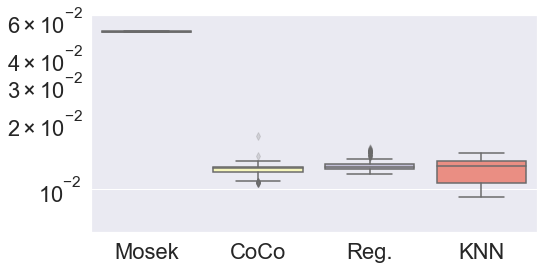}
        \vspace{-5pt}
        \caption{Solution times [s]}
        \label{fig:manipulation_time}
    \end{subfigure}
    \hfill
    \begin{subfigure}[t]{0.33\textwidth}
        \centering
        \includegraphics[height=0.7in,clip]{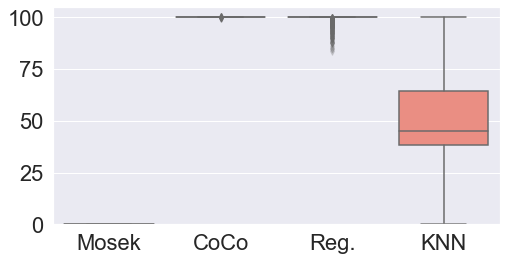}
        \vspace{-5pt}
        \caption{Normalized reward [\%]}
        \label{fig:manipulation_cost}
    \end{subfigure}
    \hfill
    \vspace{1em}
  \caption{Simulation results for manipulation example. \coco{} reduces solution times for (10) between 2–3 orders of magnitude.}
\label{fig:manipulation_results}
\vspace{-3pt}
\end{figure*}

We first study the cart-pole~\iftoggle{ext}{with wall system shown in~\cref{fig:cart-pole_soft_walls}}{with wall system}, a well-known underactuated, multi-contact problem in robot control~\cite{DeitsKoolenEtAl2019,MarcucciTedrake2020}.
The system consists of a cart and pole and the optimal control problem entails regulating the system from initial state $x_0$ to a goal $x_g$.
The non-convexity of the problem stems from four binary variables $\bin_t \in \reals^4$ introduced at each time step to enforce the logical constraint that the contact force from the wall only becomes active when the tip of the pole makes contact with either wall.
The parameter space $\theta \in \reals^{8}$ for this problem is comprised of the initial state $x_0 \in \reals^4$ and goal state $x_g \in \reals^4$.
We refer the reader to \iftoggle{ext}{the Appendix}{~\cite{CauligiCulbertsonEtAl2020}} for a full derivation of system constraints.

\subsubsection{Results}
\vspace{-2pt}
Numerical results for the cart-pole system are given in~\cref{fig:cart-pole_results}. We set the horizon $N$ to the value 10, resulting in a total of 40 binary variables. The training set consists of 90 thousand problems generated using parameters sampled from the parameter distribution $p (\Theta)$ and evaluation metrics presented for a test set of ten thousand problems. \cref{fig:cart-pole_percent_succ} reports the percent of feasible solutions found over the test set. For the commercial solver, branch-and-bound is timed out after 50ms and, for~\coco{}, ten candidate logical strategies are evaluated before the algorithm terminates with failure. Computation times shown in~\cref{fig:cart-pole_time} include the inference step to generate a candidate binary solution (\ie{} the forward pass of the network, nearest neighbor lookup, etc.) plus solution time for solving convex relaxations before a feasible solution was found. \cref{fig:cart-pole_solved} and~\cref{fig:cart-pole_cost} report the number of convex relaxations solved per problem and the cost of the feasible solution relative to the globally optimal solution, respectively. Note that that~\cref{fig:cart-pole_solved} does not include the number of convex relaxations solved by Mosek as the~\verb|cvxpy| interface does not provide this information.

We see that \coco{} outperforms the commercial solver Mosek and the two other benchmarks. As shown in~\cref{fig:cart-pole_percent_succ}, \coco{} finds feasible solutions for 99\% of the problems, compared to 82\% and 69\% for the regressor and KNN, respectively. Mosek (timed out at 50ms) finds feasible solutions for 91\% of the test set and only 44\% of these solutions correspond to the globally optimal solution. As~\crefrange{fig:cart-pole_solved}{fig:cart-pole_cost} show,~\coco{} finds the globally optimal solution after one QP solve for 98\% of its feasible solutions.

\subsection{Task-Oriented Optimization of Dexterous Grasps}
\iftoggle{ext}{
    \begin{figure}[h]
    \centering
    \includegraphics[scale=0.2,]{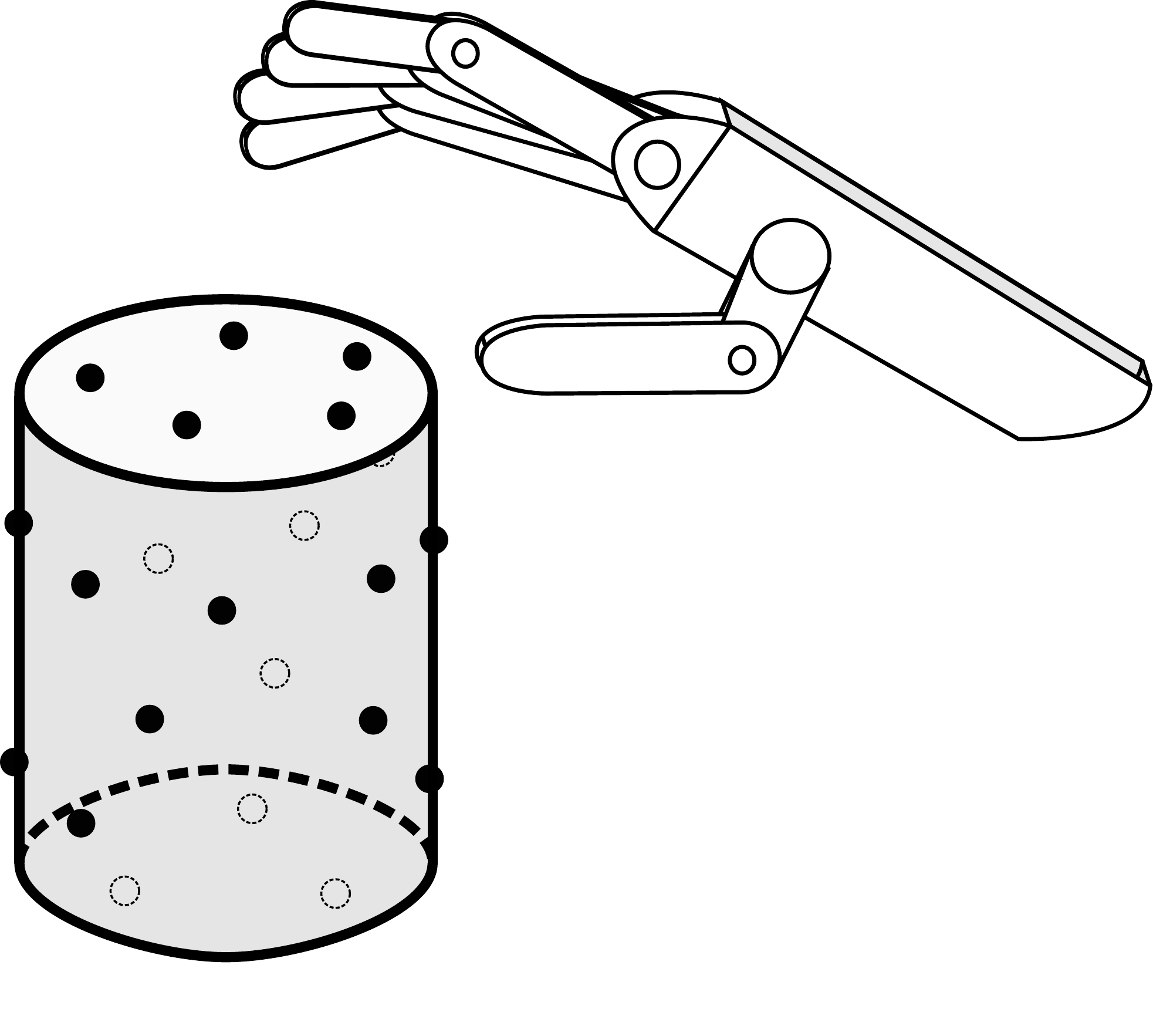}
    \caption{Schematic of dexterous grasping problem. Here, a robotic hand with $n = 5$ fingers chooses from $M$ potential contact points to optimize a task-specific grasp metric.\\}
    \label{fig:task-specific_grasping}
    \end{figure}
}{}

The next problem considered is that of grasp optimization for task-specific dexterous grasping\iftoggle{ext}{ shown in~\cref{fig:task-specific_grasping}}{}. Dexterous grasping with multi-fingered hands is a challenging problem due to both the number of contact modes that must be assigned for a stable grasp and because the resulting grasp must be able to execute the desired task under consideration. Task-agnostic grasp optimization problems generally entail solving challenging non-convex optimization problems, making them prohibitively expensive for applications~\iftoggle{ext}{replanning~\cite{FerrariCanny1992}}{replanning}.

Thus, we are interested in enabling online computation of optimal dexterous grasps for fast replanning and regrasping. Specifically, we focus our attention on the problem of task-specific grasping where the grasp sequence is chosen with a particular task such as pushing or rotating a tool in mind. Task-specific grasp optimization can be posed as an \ac{micp} where the binary variables indicate which fingers are in contact and the objective function is the grasp metric chosen to capture the quality of a grasp for a particular task~\cite{HaschkeSteilEtAl2005}.

Specifically, we consider the problem of choosing $n$ contact points for a multifingered robot hand from a set of points $p_1,\ldots,p_M \in \reals^3$, sampled from the object surface in order to optimize the task-oriented grasp metric from~\cite{HaschkeSteilEtAl2005}. The ``control'' actions optimized for are the local contact forces $f_i\in \reals^3$ such that they also satisfy \textit{friction cone} and grasp matrix constraints.
We then use binary variables $\bin$ to ensure that contact forces are not applied at all candidate points and enforce the constraint,
\vspace{-3pt}
\begin{equation*}
f_i^z \leq f^{z,\textrm{max}} \bin_i,
\vspace{-3pt}
\end{equation*}
where $f_i^z$ is the z-axis component of the local contact force.

We use the task-based grasp metric introduced in~\cite{HaschkeSteilEtAl2005} and consider \textit{task wrenches} $\hat{F}_t$, which are specific directions in wrench space that characterize the applied wrenches necessary to complete the task.
A task can then be expressed by a set of wrenches which must be generated and this set can be characterized as the positive span of $T$ task vectors.
Given a task wrench $\hat{F}_t$, let $\alpha_t$ be its associated grasp quality and $w_t \geq 0$ its task weighting.
Then, the grasp quality metric from the $T$ task vectors is given by,
\vspace{-3pt}
\begin{equation*}
\mu(\bin, \hat{F}_1, \ldots, \hat{F}_T) = \textstyle \sum \limits_{t=1}^{T} w_t \alpha_t.
\vspace{-3pt}
\end{equation*}
This grasp metric corresponds to the volume of the polyhedron defined by the vectors $w_t \alpha_t \hat{F}_t$ and can be computed by solving $T$ \acp{socp}.
Thus, this problem is a \ac{misocp} with $M$ binary variables and the parameter vector $\theta \in \reals^{12}$ consists of the desired weights for a task vector $\hat{F}_t$ which correspond to the basis vectors $\pm e_i \in \reals^6$ for $i = 1, \ldots, 6$.

\subsubsection{Results}
The numerical experiments consist of planning grasps for a four finger manipulator $n=4$. We consider a set of $M$ candidate grasp points, with $M$ equal to 30, for a single rigid body. The training set consists of 4,500 problems and the weights $w_i$, where $w_i > 0$, are generated by calculating the softmax of a vector sampled from a multivariate normal distribution with covariance matrix $\Sigma = 10\mathbf{I}$.

\cref{fig:manipulation_results} show the results for this system. As the primary point of comparison, we compare the optimality of the feasible solutions found and computation time. Indeed, we see in~\cref{fig:manipulation_cost} that \coco{} finds the globally optimal grasp for 99\% of the problems while maximizing the grasp metric is challenging for the benchmarks. As any grasp mode sequence with four contacts leads to a feasible solution for the problem, we see in~\crefrange{fig:manipulation_time}{fig:manipulation_cost} that timing out Mosek at 50ms leads to highly suboptimal solutions for this particular problem. Moreover, finding a high quality solution after solving only one \ac{socp} relaxation leads to solution times on the order of tens of milliseconds for \coco{}. Thus, \coco{} allows for high quality feasible solutions for \acp{misocp}, whereas a commercial solver would lead to highly suboptimal, low quality grasp solutions.

\subsection{Free-Flying Space Robots}
\begin{figure*}[t!]
    \centering
    \hfill
    \begin{subfigure}[t]{0.23\textwidth}
        \includegraphics[width=\textwidth]{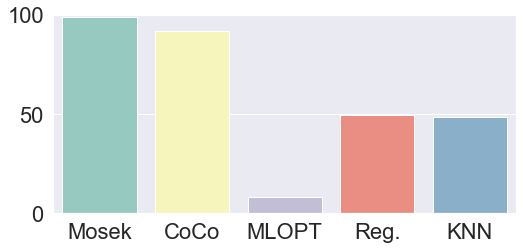}
        \caption{Success percentage}
        \label{fig:free_flyer_percent_succ}
    \end{subfigure}
    \hfill
    \begin{subfigure}[t]{0.23\textwidth}
        \includegraphics[height=0.75in,clip]{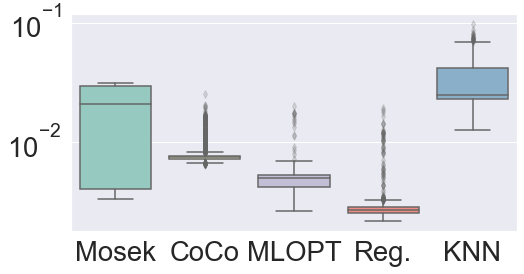}
        \caption{Solution times [s]}
        \label{fig:free_flyer_time}
    \end{subfigure}
    \hfill
    \begin{subfigure}[t]{0.23\textwidth}
        \includegraphics[height=0.75in,clip]{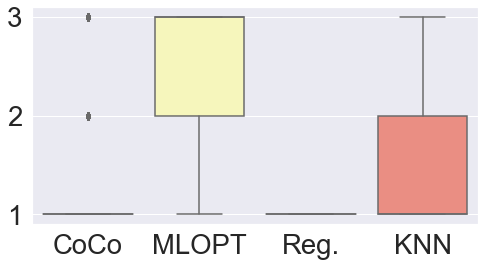}
        \caption{Num. QCQPs solved}
        \label{fig:free_flyer_solved}
    \end{subfigure}
    \hfill
    \begin{subfigure}[t]{0.23\textwidth}
        \includegraphics[width=\textwidth]{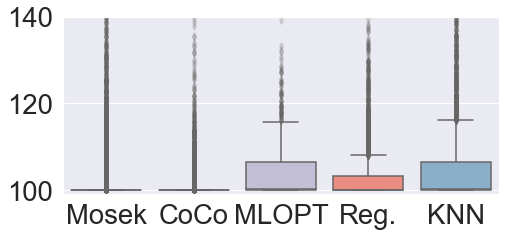}
        \caption{Normalized cost [\%]}
        \label{fig:free_flyer_cost}
    \end{subfigure}
    \hfill
    \vspace{1.5em}
  \caption{Simulation results for the free-flyer show how task-specific strategies are necessary for using \coco{}.}
\label{fig:free_flyer_results}
\vspace{0.5em}
\end{figure*}

A fundamental problem in robotics that is inherently combinatorial is that of motion planning in the presence of obstacles. Here, we study a free-flying spacecraft robot that must navigate around obstacles on a planar workspace with linear dynamics. We show how the use of task-specific logical strategies allows for (1) this problem to become tractable for application of \coco{} and (2) the learned strategy classifier $\hat{h}_\phi$ to be used at test time for \acp{micp} with a different number of binary variables $n_\bin$ than from the training set. 

The system state $x_t \in \reals^4$ consists of the position $p_t \in \reals^2$ and velocity $v_t \in \reals^2$. The planning problem is to regulate the robot from an initial state $x_0$ to a goal state $x_g$ while satisfying dynamics and actuator constraints. The crucial constraint that renders the problem non-convex is the safety constraint $x_t \in \mathcal{X}_\textrm{safe}$, as $\mathcal{X}_\textrm{safe}$ is typically a highly non-convex region of the workspace and solving the planning problem requires a global combinatorial search. For the free-flying space robot, a popular approach to solving the motion planning problem has been to pose it as an \ac{micp}~\iftoggle{ext}{\cite{LandryDeitsEtAl2016,SchouwenaarsDeMoorEtAl2001}}{\cite{LandryDeitsEtAl2016}}. In this formulation, given $N_\textrm{obs}$ obstacles, the workspace is first decomposed into keep-in and keep-out zones and binary variables $\bin$ used to enforce collision avoidance with the keep out regions.
Due to the $\ell_2$-norm constraints imposed on the thruster forces, this problem is a \ac{miqcqp} with $4N_{\mathrm{obs}}N$ binary variables. The parameters $\theta$ for this problem include the initial state $x_0$, goal state $x_g$, and position of obstacles $\{ (x^m_\textrm{min}, y^m_\textrm{min}, x^m_\textrm{max}, y^m_\textrm{max}) \}_{m=1}^{N_\textrm{obs}}$.

\subsubsection{Use of Task-Specific Logical Strategies}
We show here how task-specific strategy decompositions are necessary to efficiently solve the free-flyer motion planning problem with \coco{} and how the underlying structure of the problem can be leveraged in order to do so.

From inspection of~\eqref{eq:obstacle_avoidance_x} and~\eqref{eq:obstacle_avoidance_y}, we note that a binary variable $\bin_t^{m,i}$ appears in constraints only with the other three binary variables for obstacle $m$ at time $t$.
Thus, our key insight here is to decompose the logical strategy on a per obstacle basis.
That is, a logical strategy $\mathcal{S} (\theta;m)$ is associated with the constraint used to enforce collision avoidance with obstacle $m$ and encodes information about the binary variable assignment for that obstacle $\{ \bin_t^m \}_{t=1}^N$.
Rather than training $N_\textrm{obs}$ separate classifiers for each obstacle, we train a single classifier with $\theta$ and append an encoding to specify which obstacle logical strategy is being queried. 

Specifically, this is accomplished in this work using the architecture shown in Figure~\ref{fig:free-flyer_table}.
First, a synthetic image of the obstacles is generated using obstacle coordinates $(x^m_\textrm{min}, y^m_\textrm{min})$ and $(x^m_\textrm{max}, y^m_\textrm{max})$.
The strategy $\mathcal{S}(\theta;m)$ being queried is indicated by coloring in obstacle $m$ with a different color in the image.
A convolutional pass is then computed over the synthetic image and the output is flattened and appended with the remaining problem parameters $\theta$ before being input to a fully connected feedforward network.

For inference, a batch of input images and problem parameters are constructed, a single forward pass computed, and the strategies for the corresponding sub-formula ranked.
One issue with decomposing the strategy queries is that the full binary variable assignment $\bin$ must be reconstructed from the individual $\bin_m$.
In this work, we consider the $n_\textrm{evals}$ highest scoring logical strategy candidates for each $\mathcal{S}(\theta;m)$ and this leads to $n_\textrm{evals}^{N_\textrm{obs}}$ candidates for $\bin$.
As this can be a prohibitively large number of solution candidates, we instead randomly sample $m_\textrm{evals}$ (where $m_\textrm{evals} << n_\textrm{evals}^{N_\textrm{obs}}$) enumerations from the set of $n_\textrm{evals}^{N_\textrm{obs}}$ candidate assignments for $\bin$.
Additionally, we ensure that the $\bin_m$ corresponding to the highest scoring candidate $\mathcal{S}(\theta;m)$ for each obstacle is included in this set of $m_\textrm{evals}$ candidate binary solutions.

The results for comparing task-specific strategies used in~\coco{} are shown in~\cref{fig:free_flyer_results}, where we also include the \ac{mlopt} from~\cite{BertsimasStellato2019} as an additional benchmark.
We note immediately in~\cref{fig:free_flyer_percent_succ} that using task-specific logical strategies with~\coco{} leads to performance gains compared to the approach used for \ac{mlopt}, with 92\% feasible solutions found for~\coco{} compared to only 8\% for \ac{mlopt}.
This disparity is attributable to the fact that \coco{} encodes 458 logical strategies versus approximately 67,000 for \ac{mlopt} over the 90,000 training problems, leading to a sparser set of training labels per class for \ac{mlopt} compared to~\coco{}.
We also see in~\crefrange{fig:free_flyer_time}{fig:free_flyer_cost} that, although Mosek finds a slightly higher number of feasible solutions,~\coco{} finds the globally optimal solution for 90\% of the problems for which a feasible solution is found and at twice the speed as Mosek.
Further, we see in~\cref{fig:free_flyer_percent_succ} that the regressor and KNN benchmarks fare poorly on the logical constraints and find feasible solutions for only 49\% of the test set.

\subsubsection{Generalization}
\begin{figure}[h]
    \centering
    \includegraphics[trim={0.8cm 0.8cm 0 0.9cm},clip,width=.38\textwidth]{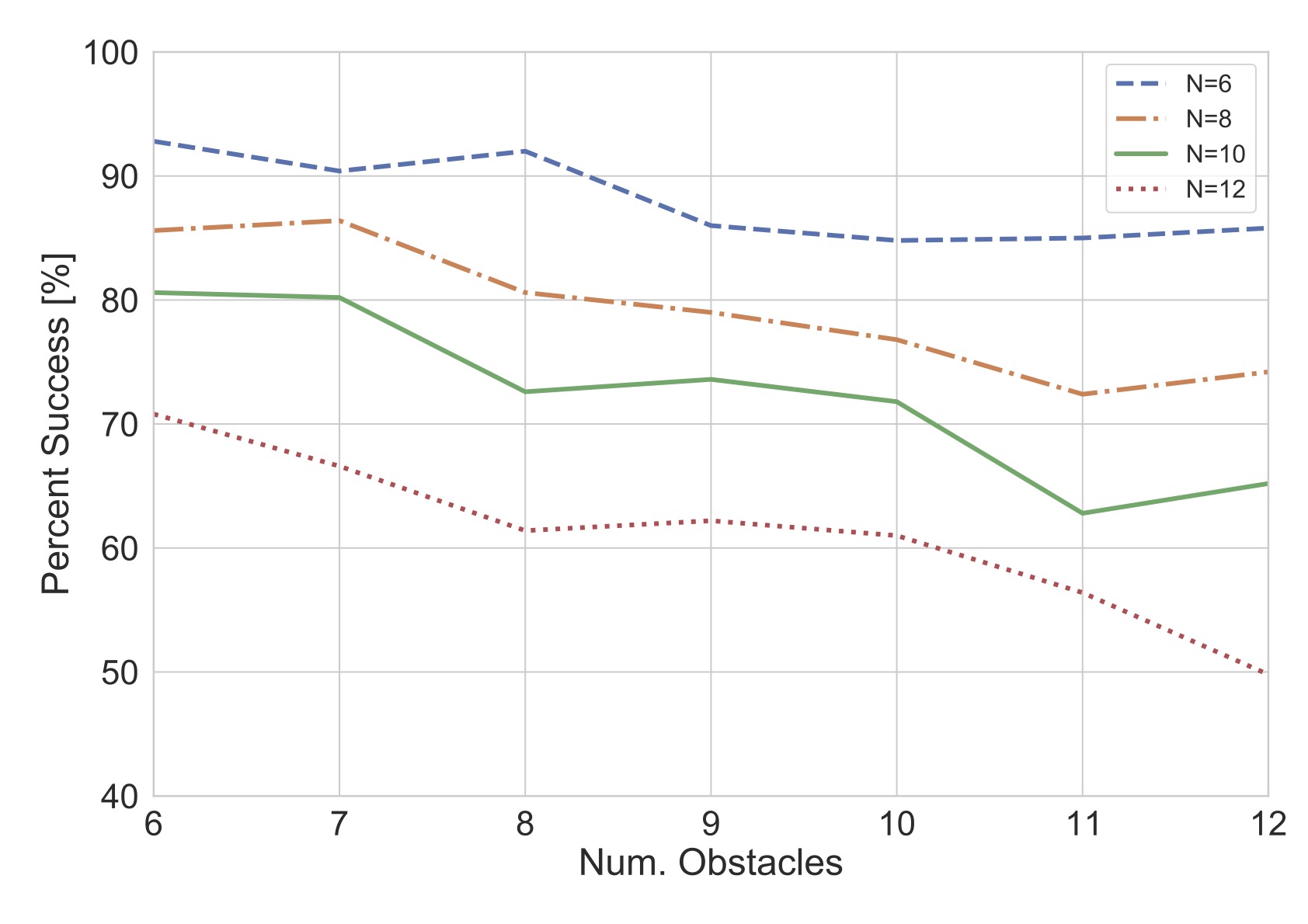}
    \caption{Generalization results for a~\coco{} strategy classifier network demonstrates roughly linear dropoff in performance as obstacles are added. The strategy classifier network was trained on an environment with $N_\mathrm{obs}=8$ for various horizons $N$.\\}
    \label{fig:coco_generalization}
    \vspace{1em}
\end{figure}

\vspace{-2pt}
One important consequence of using a convolutional pass to query the strategy sub-formula $\mathcal{S}(\theta;m)$ is that it can be used for inference in problems with a different number of $N_\textrm{obs}$ than from the training set.
Here, we evaluate the ability of~\coco{} to generalize to a distribution of problems with a varying number of binary obstacles.
We train multiple strategy classifiers corresponding to horizons of $N=\{5,7,9,11\}$ with environments of eight obstacles.
\cref{fig:coco_generalization} shows the performance of applying these networks in solving problems with an increasing number of obstacles $N_\mathrm{obs}= \{6, \ldots, 12\}$.
As shown, we see that the efficacy of the strategy classifier diminishes with an increasing horizon length $N$ due to a corresponding increase in the number of strategies.
Intuitively, we also see that performance decreases with an increase in obstacles simply due to the increased difficulty of the planning problem.
However, we note that the performance dropoff remains roughly linear rather than an exponential decrease of performance stemming from including additional binary variables.
Thus, given knowledge of the robot operating environment, the use of a trained classifier can be limited to scenarios in which the number of obstacles does not lead to a dramatic dropoff in performance.

\subsubsection{Timeout Performance}
\begin{figure}[t]
\centering
\includegraphics[clip,width=0.75\columnwidth]{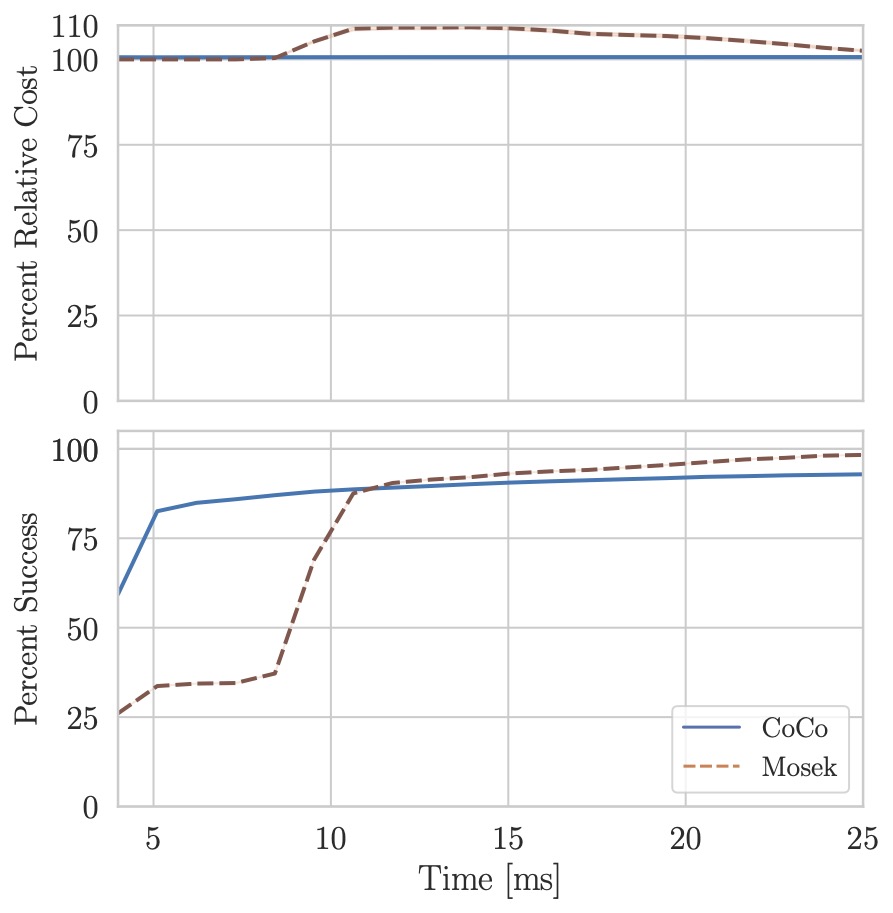}
\caption{ We compare the performance of~\coco{} (blue) and Mosek (brown) after timing out both on a test set of~\acp{miqcqp}. We see in the bottom plot that timing out Mosek leads to a reduced number of feasible solutions found compared with~\coco{}.
As the cutoff time increases, we see in the top plot that if Mosek finds a feasible solution, then these feasible solutions are suboptimal compared to~\coco{}, which generally finds the globally optimal solution even when timed out early.\\}
\label{fig:free_flyer_cutoff}
\vspace{5pt}
\end{figure}
Here, we compare~\coco{} with a commercial solver that is timed out with
\iftoggle{ext}{a prespecified termination time (\ie{}, the incumbent solution from branch-and-bound is returned).}{a prespecified termination time.}
For~\coco{}, we terminate~\coco{} either when a feasible solution is found or after the termination time has been exceeded. \cref{fig:free_flyer_cutoff} compares the percent of feasible solutions found between~\coco{} and Mosek.
We see that~\coco{} finds a feasible solution for the majority of problems within about
\iftoggle{ext}{five milliseconds, which is approximately the time required to compute a forward pass of the CNN and solve a single convex relaxation.}{five milliseconds.}
However, Mosek requires twice the computation time before it finds a feasible solution for the majority of problems.\\
\vspace{-2pt}
Moreover, we see that the solutions found by~\coco{} are effectively the globally optimal solution for that problem, whereas the Mosek's  incumbent solution from branch-and-bound is often suboptimal until branch-and-bound terminates.
Thus, in tasks where Mosek is allowed to run its full course, Mosek will indeed find the globally optimal solution, but a designer can weigh the tradeoffs between quickly finding a high quality feasible solution using~\coco{} or allowing Mosek to terminate.
Finally, we further note that in applications requiring a certificate of optimality, a feasible solution found by~\coco{} can be used as the incumbent and provide a tighter upper bound for branch-and-bound.

\section{Conclusion}
\vspace{-5pt}
In this work, we presented~\coco{}, a data-driven framework to find high quality feasible solutions for \acp{micp} used in robot planning and control problems.
We demonstrated how problem structure arising in robot tasks can be utilized effectively in a supervised learning framework.
Specifically, we introduced the notion of task-relevant logical strategies to exploit such problem structure and showed how they improve the performance of the trained strategy classifier.
We showed through numerical experiments that~\coco{} improves solution speeds by 1-2 orders of magnitude with only a slight loss of optimality, compared to low-quality feasible solutions found by the commercial solver and benchmark data-driven approaches.
Finally, we showed how \coco{} can uniquely be used to solve problems with a varying number of discrete decision variables and how this allows for solving a new set of tasks online.
To this end, we believe that a promising direction of work is to improve the classifier performance given new tasks online.
One future approach could be to explore meta-learning to allow for~\coco{} to adapt network parameters online for improved performance using information gained from solving problems online.

{\small
\renewcommand{\baselinestretch}{0.92}
\bibliographystyle{IEEEtran}
\bibliography{ASL_papers,main}}

\iftoggle{ext}{\section*{Appendix}
\label{sec:appendix}
We review the system dynamics and constraints for each~\ac{micp} studied in this work.

\subsection{Cart-Pole with Soft Walls}
As depicted in \cref{fig:cart-pole_soft_walls}, the system consists of a cart and pole and the optimal control problem entails regulating the system to a goal $x_g$:
\begin{equation} \label{eq:cart-pole_ocp}
\begin{array}{ll}
\underset{x_{0:N},,u_{0:N-1},\bin}{\textrm{minimize}} & \displaystyle \|x_N-x_g\|_{2} + \sum \limits_{\tau=0}^{N-1} \|x_\tau-x_g\|_{2} + \|u_\tau\|_{2} \\
\text{subject to} & x_{t+1} = Ax_t + Bu_t + G s_t,\quad t = 0,\mydots,N-1 \\
& u_{\textrm{min}} \leq u_t \leq u_{\textrm{max}}, \quad t = 0,\mydots,N-1 \\
& s_t = \begin{cases}
    \kappa\lambda_t + \nu \gamma_t & \text{if } \lambda_t\geq0 \text{ and } \kappa\lambda_t+\nu\gamma_t \geq 0\\
    0,              & \text{otherwise}
\end{cases} \quad\\
& \qquad\qquad\qquad t = 0,\mydots,N-1\\
& x_{\textrm{min}} \leq x_t \leq x_{\textrm{max}}, \quad t = 0,\mydots,N \\
& x_0 = x_{\textrm{init}}\\
& \bin \in \left\{0,1\right\}^{4\times (N-1)},
\end{array}
\end{equation}
where the state $x_t \in \reals^4$ consists of the position of the cart $x_t^1$, angle of the pole $x_t^2$, and their derivatives $x_t^3$ and $x_t^4$, respectively. The force applied to the cart is $u_t\in\reals$ and $s_t\in\reals^2$ are the contact forces imparted by the two walls. The relative distance of the tip of the pole with respect to the left and right walls is $\lambda_t \in \reals^2$ and the time derivative of this relative distance $\gamma_t \in \reals^2$. Finally, $\kappa$ and $\nu$ are parameters associated with the soft contact model used.

As the contact force $s_t$ becomes active only when the tip of the pole makes contact with either wall, we must introduce binary variables to enforce the logical constraints given in \eqref{eq:cart-pole_ocp}. We denote the relative distance of the tip of the pole with respect to the left and right walls as $\lambda_{t}^1$ and $\lambda_{t}^2$, respectively:
\begin{equation}
\begin{array}{l}
\lambda_t^1 = -x_t^1 + \ell x_t^2 - d \nonumber\\
\lambda_t^2 = x_t^1 - \ell x_t^2 - d, \nonumber
\end{array}
\end{equation}
where $\ell$ is the length of the pole and $d$ half the distance between the walls. The time derivatives of $\lambda_t^1$ and $\lambda_t^2$ are
\begin{equation}
\begin{array}{l}
\gamma_t^1 = -x_t^3 + \ell x_t^4 \nonumber\\
\gamma_t^2 = x_t^3 - \ell x_t^4. \nonumber
\end{array}
\end{equation}

To constrain contact forces $s_t$ to become active only when the pole tip strikes a wall, we introduce four binary variables $\bin_t^i, \quad i = 1,\mydots,4$. Using the formulation from \cite{BemporadMorari1999}, we enforce the following constraints for $k = 1, 2$:
\begin{equation} \label{eq:cartpole_logic}
\begin{array}{l}
\lambda_\textrm{min}^k(1-\bin_t^{(2k-1)}) \leq \lambda_t^k \leq \lambda_\textrm{max}^k\bin_t^{(2k-1)} \nonumber\\
s_\textrm{min}^k(1-\bin_t^{(2k)}) \leq \kappa \lambda_t^k + \nu \gamma_t^k \leq s_\textrm{max}^k\bin_t^{(2k)} \nonumber \\
\end{array}
\end{equation}
Finally, we impose constraints on $s_t$, for $k=1,2$:
\begin{equation} \label{eq:cartpole_contact}
\begin{array}{l}
\nu \gamma_\textrm{max}^{k}(\bin_t^{(2k-1)}-1) \leq s_t^{k} - \kappa \lambda_t^{k} - \nu \gamma_t^{k} \leq s_\textrm{min}^k(\bin_t^{(2k)}-1) \nonumber \\
\end{array}
\end{equation}
There are then a total of $4N$ integer variables in this MIQP.

\subsection{Task-Oriented Optimization of Dexterous Grasps}
The task-specific dexterous grasping problem entails choosing $n$ contact points out of $M$ in order to optimize the task-oriented grasp metric from~\cite{HaschkeSteilEtAl2005}. Each contact point $p_i \in \reals^3$ is sampled from the object surface and contacts between the finger and the object are modeled as point contacts with friction. The force that can be applied by each finger in the local contact force is $f_i = (f_i^x, f_i^y, f_i^z)\in \reals^3$, where the local coordinate frame has the $x$- and $y$-axes tangent to the surface, and the $z$-axis along the inward surface normal. Intuitively, $f_i^z$ is the component of the contact force which is normal to the object surface, and $f_i^x, f_i^y$ are its tangential components. 

Under this contact model, we constrain the contact force $f_i$ to lie within the \textit{friction cone} $\mathcal{K}^{(i)}$. Let us further define the contact force vector $f = (f_1, \ldots, f_M) \in \reals^{3M}$, which is the vector of all contact forces.

Using the definition of a grasp matrix from \cite{FerrariCanny1992}, we can express the wrench applied to the object (from all contact forces) as $G f$, where $G = \left[G_1, \ldots, G_M\right]$.

However, contact forces may not be applied at all candidate points. To this end, we introduce the logical variables $\bin_i \in \left\{0,1\right\}$, with $\bin_i= 1$ iff point $p_i$ is selected for the grasp. Thus, we enforce the constraint $$f_i^z \leq \bin_i,$$ which constrains the normal forces of all unused grasps to be zero, and to be bounded by unity otherwise. Thus, for a choice of grasps $\bin = (\bin_1, \mydots, \bin_M)$, the set of possible object wrenches is defined as $\mathcal{W}(\bin) = \left\{G f \mid f\in \mathcal{K}_i, f_i^z \leq \bin_i \right\}.$

In \cite{HaschkeSteilEtAl2005}, the authors propose a task-based grasp metric using \textit{task wrenches} $\hat{F}_t$, which are specific directions in wrench space that characterize the applied wrenches necessary to complete the task. For instance, if the desired task is to push the object along the $+x$-axis, then this task could be described using $\hat{F} = (1,0,0,0,0,0)$, and so on. For a task described by a single wrench, the grasp quality can be defined as 
\begin{equation*}
    \mu_1(\bin, \hat{F}_t) = \sup \left\{\alpha \geq 0 \mid  \alpha \hat{F}_t \in \partial \mathcal{W}(\bin) \right\},
\end{equation*}
where $\partial \mathcal{W}$ denotes the boundary of $\mathcal{W}.$

However, most tasks are best described by a set of wrenches which must be generated, rather than a single direction in wrench space. Thus, the authors propose describing this set as the positive span of $T$ task vectors; in turn, the grasp metric is defined as $$\mu(\bin, \hat{F}_1, \ldots, \hat{F}_T) = \textstyle \sum \limits_{t=1}^{T} w_t \alpha_t,$$ where $w_i \geq 0$ are the relative weightings of the task vectors, and $\alpha_t = \mu_1(\bin,\hat{F}_t).$ This can, in turn, be computed by solving $T$~\acp{socp}.

We seek $\bin^*$ which maximizes this grasp metric, which yields a~\ac{misocp}:
\begin{equation} \label{eq:manip_prob}
\begin{array}{ll}
\underset{\alpha,f_{1:M},\bin}{\textrm{maximize}} & \displaystyle \sum\limits_{t=1}^{T} w_t \alpha_t\\
\text{subject to} &G f^t = \alpha_t \hat{F}_t , \quad t = 1, \ldots, T\\
 &f_i^t \in \mathcal{K}^{(i)}, \quad i = 1, \ldots M,\; t = 1, \ldots, T\\
 & f_i^{z,t} \leq \bin_i, \quad i = 1, \ldots, M,\; t = 1, \ldots, T\\
 & \sum\limits_{i=1}^M \bin_i \leq n\\
 & \bin \in \left\{0,1\right\}^M
\end{array}
\end{equation}

\subsection{Free-Flying Space Robots}
We let $p_t \in \reals^2$ be the robot position and $v_t \in \reals^2$ the velocity in the 2-dimensional plane. The robot state is $x_t = (p_t, v_t)$ and the input $u_t\in\reals^2$ consists of the forces produced by the thruster. Letting $\mathcal{X}_{\textrm{safe}}$ be the free space which the robot must navigate through, the optimal control problem is to plan a collision free trajectory towards a goal state $x_g$:
\begin{equation} \label{eq:free-flyer_ocp}
\begin{array}{ll}
\underset{x_{0:N},,u_{0:N-1},\bin}{\textrm{minimize}} & \displaystyle \|x_N-x_g\|_{2} + \sum \limits_{\tau=0}^{N-1} \|x_\tau-x_g\|_{2} + \|u_\tau\|_{2} \\
\text{subject to} & x_{t+1} = Ax_t + Bu_t, \!\! \quad t = 0,\mydots,N-1 \\
& ||u_t||_{2} \leq u_{\textrm{max}},\!\! \quad t = 0,\mydots,N-1 \\
& x_{\textrm{min}} \leq x_t \leq x_{\textrm{max}},\!\! \quad t = 0,\mydots,N \\
& x_0 = x_{\textrm{init}}\\
& x_t \in \mathcal{X}_{\textrm{safe}},\!\! \quad t=0,\mydots,N\\
& \bin \in \left\{0,1\right\}^{4N_\textrm{obs}\times N}.
\end{array}
\end{equation}

The constraint $x_t \in \mathcal{X}_{\textrm{safe}}$ is the primary constraint of interest as it renders the problem non-convex. In this work, we consider axis-aligned rectangular obstacles. An obstacle $m$ is parametrized by the coordinates of its lower-left hand corner $(x^m_\textrm{min}, y^m_\textrm{min})$ and upper right-hand corner $(x^m_\textrm{max}, y^m_\textrm{max})$. Given the state $x_t$, the collision avoidance constraints with respect to obstacle $m$ at time $t$ are:
\begin{align}
x_{\textrm{max}}^m - M\bin_{t}^{m,1} \leq x_t^1 \leq x_{\textrm{min}}^m + M\bin_{t}^{m,2} \label{eq:obstacle_avoidance_x}\\
y_{\textrm{max}}^m - M\bin_{t}^{m,3} \leq x_t^2 \leq y_{\textrm{min}}^m + M\bin_{t}^{m,4} \label{eq:obstacle_avoidance_y}
\end{align}}{}

\end{document}